# THE CONTROLLABILITY OF PLANNING, RESPONSIBILITY, AND SECURITY IN AUTOMATIC DRIVING TECHNOLOGY

WanDan[1#], ZhanHao[2#*]
[1] Institute for AI Moral Decision-Making
Hunan Normal University, ChangSha, China
13635464@qq.com

[2] Institute for AI Moral Decision-Making
Hunan Normal University, ChangSha, China
zhanhao@smail.hunnu.edu.cn

## ABSTRACT

People hope automated driving technology is always in a stable and controllable state; specifically, it can be divided into controllable planning, controllable responsibility, and controllable information. When this controllability is undermined, it brings about the problems, e.g., trolley dilemma, responsibility attribution, information leakage, and security. This article discusses these three types of issues separately and clarifies the misunderstandings.

[*] Corresponding Author.

[#] These authors contributed equally to the work.

## 1  INTRODUCTION

Automated driving technology has become the focus of research institutions and manufacturers in recent years. Both traditional automakers and Internet companies have long been involved in the development of automated driving technology and have achieved certain results. In 2017, GM equipped the Super Cruise automatic driving function on the Cadillac CT6. In April of the same year, Baidu released the Apollo self-driving vehicles platform. In July, Audi officially released the Audi A8, and its automated driving system Traffic Jam Pilot reached Level 3. In October, Waymo completed the first social road test of Level 4 self-driving vehicles for the first time. In April 2018, Baidu launched the test ride of Level 4 Baidu driverless bus "Apolon," and announced the automated driving bus entered the mass production phase in July.

The rapid development of automated driving technology has also led to a lot of discussions - most of which are concerned about the widespread use of automated driving technology. Some people believe that there are many ethical and legal dilemmas in automated driving technology, and they must be constrained to meet people's needs before they are applied to real life.

It is hoped that automated driving technology will always be in a controlled state while reducing the driving accident rate. It means that we do not impose strict restrictions on automated driving technology. While recognizing its shortcomings, we effectively recognize and control the cost of using this technology, so that it is in a state of controllable balance.

The controllability of automatic driving system is reflected in three points:

　　1. Controllable Planning, and its representative problem is the trolley problem;

　　2. Controllable Responsibility, and its representative problem is responsibility subject;

　　3. Controllable Security, and its representative issues are information leakage and information security.

People's doubts about automated driving technology are mainly concentrated in these three parts. From another point of view, as long as the planning, responsibility and information control of the automated driving technology are realized, the doubts about the automated driving technology are solved to some extent.

## 2  CONTROLLABLE PLANNING: TROLLEY PROBLEM UNDER THE AUTOMATIC DRIVING SCENE

Planning is the decision and command issued by the automated driving system to the driverless car, that is, the path planning and manipulation instructions, such as "what kind of driving path is legal" or "when faced with an inevitable accident, what kind of decision is ethical" are all questions about whether the automatic driving system can truly achieve "controllable planning". Among them, the most typical martyrdom is the Trolley Problem.

Trolley Problem was initially proposed by Philippa Foot (1967, pp.152-161): a driver driving a "runaway tram" on a forked track, with five people on the original route tied to the track. If the trolley continues to drive, then the five will die. And on the other road, there is only one person related to the track. The driver is faced with two choices: either doing nothing so that the tram will hit five people; or turn to another road so that only one person must be sacrificed. What would you do if you are a driver?

However, people have neglected (1) if one faced to the trolley problem, he/she should accept the losses after his/her any decision-making; (2) for the automatic system, if you want to make decisions under the trolley problem, you need to set up a program to deal with it. So it means the automatic driving system must be designed with the loss more or less. In other words, the automated driving system will make decisions to sacrifice humanity in a certain situation inevitably, which is difficult to accept for us.

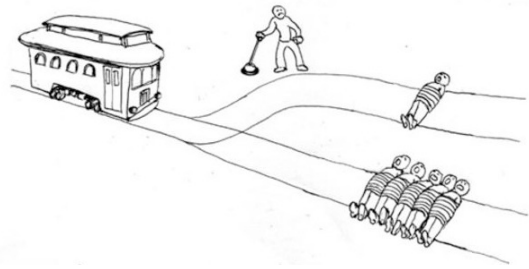

**Figure 1: Foot`s trolley problem. People have to choose between killing one and killing five.**

However, people have neglected (1) if one faced to the trolley problem, he/she should accept the losses after his/her any decision-making; (2) for the automatic system, if you want to make decisions under the trolley problem, you need to set up a program to deal with it. So it means the automatic driving system must be designed with the loss more or less. In other words, the automated driving system will make decisions to sacrifice humanity in a certain situation inevitably, which is difficult to accept for us.

In fact, as early as 2016, Mercedes-Benz said in public that the automatic driving system might give priority to protecting the safety of passengers in the car under the premise of damage, and sacrifice the outside people, which was strongly resisted by the public. People think that the automatic driving system has no right to make choices about life. On the other hand, as the official guidance document for the ethical code of self-driving vehicles, the Ethics Commission Automated and Connected Driving Report issued by the Federal Minister of German Transport and Digital Infrastructure also explicitly mentions that the autopilot system with "loss prediction" is not allowed.

The third ruler in the Report says, "the public sector is responsible for guaranteeing the safety of the automated and connected systems introduced and licensed in the public street environment," and "the guiding principle is the avoidance of accidents." At the same time, the fifth ruler makes it more apparent that the priority in the trolley problem is to avoid accidents as much as possible, "Based on state of the art, the technology must be designed in such a way that critical situations do not arise in the first place. These include dilemma situations, in other words, a situation in which an automated vehicle has to "decide" which of two evils, between which there can be no trade-off, it necessarily has to perform."

It is unacceptable for public and government that automatic driving system makes decisions with pre-set of loss, as evidenced by these examples and guidelines. Furthermore, the government has tried to exclude the default of losses from the systems.

Once "loss prediction" is rejected into the automatic driving system, the tram dilemma under the automated driving based on the lossy preset will no longer exist, and our doubts about the "controllable planning" will be resolved.

## 3    CONTROLLABLE RESPONSIBILITY: FROM RESPONSIBILITY TO BUSINESS BEHAVIOUR

Responsibility attribution problem is another widely discussed issue in automated driving technology: how should the liability of the accident be determined if an accident occurs during automated driving?

In the automation grading system introduced by the Society of Automotive Engineers (SAE), automatic driving vehicles are divided into six phases:

Level 0, No Automation, means the driver has absolute control during the driving process;

Level 1, Driver Assistance, means the driver can transfer part of the control to the system, under the limited time and limited conditions;

Level 2, Partial Automation, means the driver can transfer all control to the system, under the limited time and limited conditions;

Level 3, Conditional Automation, means the driver gives the appropriate response when the emergency occurs, and the system completes most of the control.

Level 4, High Automation, means the driver does not need to control in some driving modes, the system can complete all control.

Level 5, Full Automation, means the driver does not need to control in all driving modes, the system can complete all control.

Level 0, Level 1, and Level 2 belong to the assisted driving phase, the automatic driving system only supports human driving, but cannot perform individual driving behavior. In Level 3, the system can independently complete most driving operations, but when the emergency occurs, the driver needs to take over. In Level 4, the system could complete all driving operations in some scenarios. In Level 5, the automatic vehicles can complete the driving operation in any scene.

| SAE level | Name | Narrative Definition | Execution of Steering and Acceleration/ Deceleration | Monitoring of Driving Environment | Fallback Performance of Dynamic Driving Task | System Capability (Driving Modes) |
|---|---|---|---|---|---|---|
| *Human driver* monitors the driving environment | | | | | | |
| 0 | No Automation | the full-time performance by the *human driver* of all aspects of the *dynamic driving task*, even when enhanced by warning or intervention systems | Human driver | Human driver | Human driver | n/a |
| 1 | Driver Assistance | the *driving mode*-specific execution by a driver assistance system of either steering or acceleration/deceleration using information about the driving environment and with the expectation that the *human driver* perform all remaining aspects of the *dynamic driving task* | Human driver and system | Human driver | Human driver | Some driving modes |
| 2 | Partial Automation | the *driving mode*-specific execution by one or more driver assistance systems of both steering and acceleration/ deceleration using information about the driving environment and with the expectation that the *human driver* perform all remaining aspects of the *dynamic driving task* | System | Human driver | Human driver | Some driving modes |
| *Automated driving system* ("system") monitors the driving environment | | | | | | |
| 3 | Conditional Automation | the *driving mode*-specific performance by an *automated driving system* of all aspects of the dynamic driving task with the expectation that the *human driver* will respond appropriately to a *request to intervene* | System | System | Human driver | Some driving modes |
| 4 | High Automation | the *driving mode*-specific performance by an automated driving system of all aspects of the *dynamic driving task*, even if a *human driver* does not respond appropriately to a *request to intervene* | System | System | System | Some driving modes |
| 5 | Full Automation | the full-time performance by an *automated driving system* of all aspects of the *dynamic driving task* under all roadway and environmental conditions that can be managed by a *human driver* | System | System | System | All driving modes |

**Figure 2: Levels of driving automation for on-road vehicles in SAE International J3016.**

In L0, L1, and L2, the automatic driving technology only plays an auxiliary role, division of responsibility is the same as traditional driving method, so there is no need to discuss the attribution of responsibility; in the L4, although all the automatic driving techniques have been implemented by automated driving, the whole process must be in a specific scene, such as freight passages and power plants, and the accident responsibility is easily defined clearly (Si Xiao, Cao Jianfeng, 2017, pp.166-173). As for the L5, it is only an imaginary; the key to the responsibility allocation problem lies in L3.

The difficulty we face is that the automatic driving system is formed by multiple agents (automakers, intelligent software providers, navigation and location providers, and others), it is extremely difficult to determine the responsibility subject in such a situation (Chen Xiaolin, 2016, pp.124-131). On the other hand, determining the source of interference is another problem because of the vulnerable system (Li Yiran, Zhou Jie, 2010, pp.156-160). As for the situation of accidents and emergency, it is even harder to distinguish responsibility subject (Chen Hui, Xu Jianbo, 2014, pp.64-70).

Scholars have tried to come up with various solutions, and some researchers believe that the criminal responsibility should be excluded if we can prove the automated vehicle is causing damage within its automatic driving range, and the current technical level can't predict and prevent the damage situation (Long Min 2018, pp.78-83). Some scholars have pointed out that the responsibility of the company can be analyzed by judging the possibility of avoidance, the decision-making ability of the driving system, and the supervision responsibility of enterprises and personnel who are manufacturing, producing, and programming (Cheng Long 2018, pp.83-89). Other researchers believe that if the development of self-driving vehicles is necessary, there will be a moral obligation to make the whole society responsible for the development of self-driving vehicles. Once the responsibility caused by self-driving needs to be assumed by one subject, the entire society will take it in the form of social insurance (Hevelke, Nidarümelin, 2015, pp. 619-630).

However, these schemes are either not accepted by the public or do not give actual solutions, so they are not able to properly resolve the responsibility of automated technology in L3. From the perspective of supporting the development of automated driving, we proposed two solutions: 1) put the Responsibility on the drivers and 2) technology level spanning.

> Option 1: put the responsibility on the drivers
>
> This option aims at driver's freedom to choose and merchandise attributes of the self-driving vehicles.  On the one hand, drivers choose their cars or automatic driving system by their free wills, and they should be responsible for those choices, so we can attribute the responsibility of the automated driving accident to the drivers if there are only two responsibility subjects, driver and self-driving vehicles.
>
> Option 2: technology level spanning
>
> This option completely solves the responsibility problem of L3 by forbidding cross-driving between self-driving vehicles and human-driving vehicles. Volvo, Ford, and other companies have claimed that they will abandon the automatic driving technology research and skip L3 to L4 or L5, because of the responsibility issues in L3. Similarly, many scholars proposed that companies should skip L3 in view of the liability problem of L3 at the Netease 2017 Future Technology Summit.

## 4   CONTROLLABLE SECURITY: INFORMATION LEAKAGE AND CONNECTED ATTACK

The last controllable obstacle comes from the online state of automated driving. Like most such states, it faces to information leakage problem and the connected security problem.

The first one means there is a risk of privacy violation and data leakage when automatic-driving vehicles have to continuously collect relevant information and data (Jiang Su, 2018,pp.180-189). But, is this risk unique to automated driving?

On the one hand, from the technical view, self-driving vehicles are no different from other digitizing equipment, in terms of information leakage. So there is no technical specificity.

On the other hand, from the content view, the type of data self-driving vehicles collected is the same as digitizing equipment. Maybe there are some objections, ordinary equipment can only collect essential data, like location, time, but no speed, rapid acceleration et al. Of course, in some degrees there is the difference between the two. However, this difference is increasingly eroded to zero with the update of digitizing equipment. For example, the APP "OKDrive" has realized to acquire the data of location, history, speed, rapid acceleration, number of sudden braking, etc. through the mobile phone GPS and sensor modules data.

Therefore, there is no automated driving system`s privacy problem, but internet information's privacy problem.

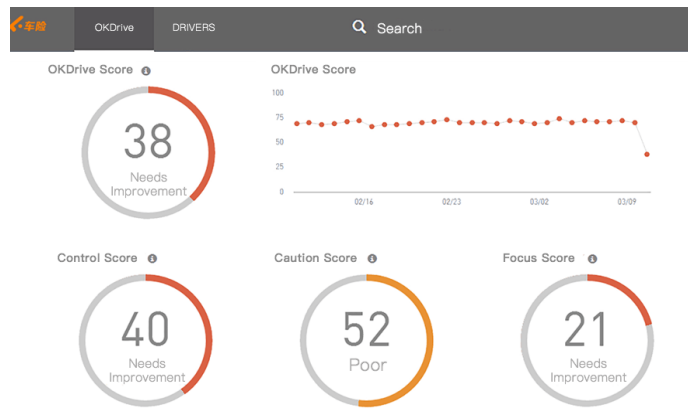

**Figure 3: OKDrive APP can track and compile data on users' driving habits, using existing smartphone functions without the need for any external hardware.**

The second one mainly refers to the risk of attacked or maliciously invaded in automated technology by the network. Unlike leakage, the security problem has their particularities: The safety problem of traditional vehicles has the following characteristics: 1. physical contact; 2. the number of the vehicles be damaged limited in a short time in traffic accidence; 3. the attack is always easy to be detected. While for the self-driving vehicles: 1. No physical contact; 2. multiple vehicles can be destroyed in a short time; 3. difficult to be detected.

The "no physical contact" has two situations, one is to invade and manipulate in the process of information transmission, such as tampering with information and sending the error information; or invade the CPU or the controller of the self-driving vehicles, then control the target or make it paralyzed, lastly hijacking the target (Gao Zhaoming, Gao Hao, 2014, pp.77-83).

Another one is to mislead self-driving cars into catastrophic accidents by providing them with false information from attackers. For example, special obstacles that can absorb radar and laser waves are used, similar to invisibility cloaks; or interfere with the perception system of the self-driving car, give the wrong information to it. In fact，as early as 2016, some scholars proved that we are able to perform jamming and spoofing attacks whit off-the-shelf hardware, which caused the Tesla's blindness and malfunction, all of which could potentially lead to crashes and impair the safety of self-driving cars. In 2017，a team of researchers from four American universities has provided a troubling preview of how self-driving cars could be tricked into making dangerous mistakes. We can disguise the stop mark as the speed limit mark, with a success rate as high as 66.67%~100%, with  some technologies, such as Camouflage Poster, Camouflage Poster Right Turn, Camouflage Graffiti, Camouflage Art (LISA-CNN), Camouflage Art (GTSRB-CNN)（Eykholt, Evtimov, Fernandes, et al., 2017, p.1631）.

The "multiple cars can be destroyed in a short time" means this destruction of self-driving cars can be done on multiple objects at the same time because of the connected state. So, the consequences will be unimaginable once self-driving cars are controlled by criminals or terrorists maliciously, considering their operating environment, mobility, and kinetic energy.

The "difficult to be detected" means that such risk often exists in a hidden form that is hard to be discovered by sense organs, which not only increase the difficulty of foreseeing and evaluation but also puts forward high demands on our defensive measures.

Fortunately, researchers have recognized the damage and severity of the connected attacks and have begun to try to come up with solutions. For example, we can build a neural network with long short-term memory (LSTM) structure to defends the man-in-the-middle and eavesdropping attacks (Ferdowsiand Saad, 2019, pp. 1371-1387)

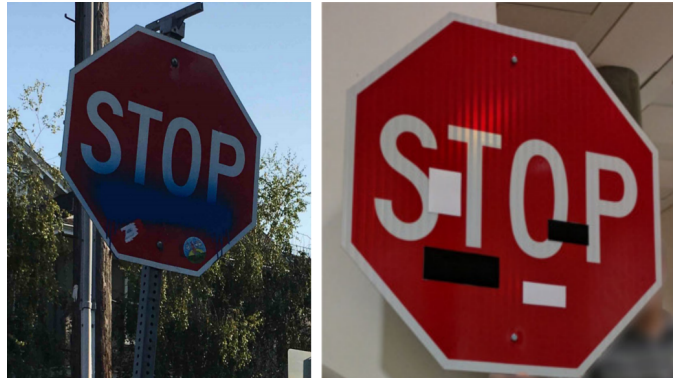

**Figure 4: The left image shows real graffiti on a Stop sign, something that most humans would not think is suspicious. The right image shows our a physical perturbation applied to a Stop sign, and this physical perturbation can trick the car into believing Stop sign really means Speed Limit 45 or something else.**

Adversarial Attack cannot be defended by weight regularization, dropout and model ensemble. There are two types of defense:

(1) Passive defense: finding the attached image without modifying the model.
   Passive defense is not to change the model itself, but to add a simple filter (e.g. smoothing filter). Feature squeeze and model B are representatives of this path, the first one reducing the color bit depth of each pixel and spatial smoothing (Weilin Xu, David Evans, Yanjun Qi, 2017), and the second one using two randomization operations: random resizing, which resizes the input images to a random size, and random padding, which pads zeros around the input images in a random manner (Cihang Xie, Jianyu Wang, Zhishuai Zhang, et al., 2017).

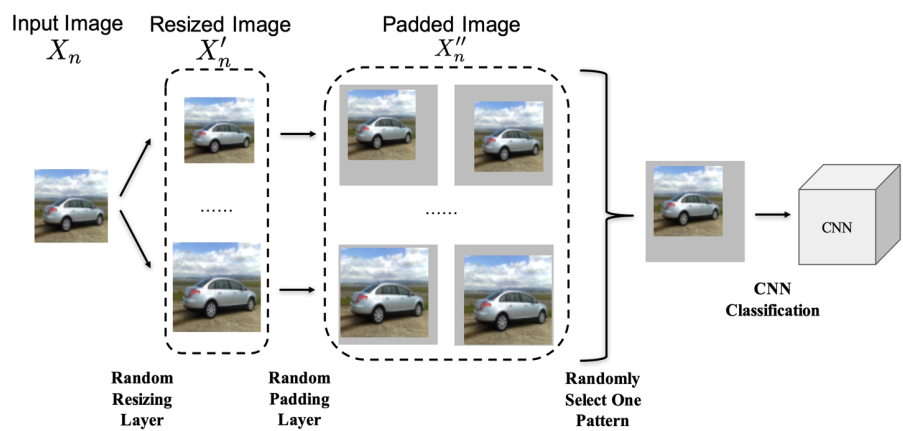

**Figure 5: The pipeline of randomization-based defense mechanism**

(2) Proactive defense: training a model that is robust to adversarial attack.

Proactive defense refers to the practice of finding and fixing flaws in a model while training it. In this way, we can find adversarial input data by an attack algorithm, but the point is what the attack algorithm is?

## 5 CONCLUSION

For a variety of reasons, misunderstandings always exist in our cognition of autonomous driving systems. Indeed, there are many problems caused by these misunderstanding, and that is why we should clarify them so that focus on the real problems.

In this paper, we discussed the trolley problem, and found that due to the existence of loss presupposition, we did not need to discuss such problems in the autonomous driving scenario; Secondly, we discussed the controllable responsibility problem, and we have programmed two solutions: 1) put the responsibility on the drivers and 2) technology level spanning. Finally, we discussed the controllable security problems, which include the information leakage problem and the connected security problem. In the part of information manufacture, we believed that there is no automated driving system's privacy problem, but Internet information's a privacy problem, and in the part of the security problem, we have proposed passive defense solution and proactive defense solution.

## 6 REFERENCES

[1]    **Bonnefon, J. F., Shariff, A., and Rahwan, I.** 2016, The Social Dilemma of Autonomous Vehicles, *Science*, 352(6293), pp.1573-1576.

[2]    **Bonnefon, J. F., Shariff, A., and Rahwan, I.** 2015, Autonomous vehicles need experimental ethics: Are we ready for utilitarian cars, *arXiv preprint*, 1510(03346).

[3]    **Eykholt, K., Evtimov, I., Fernandes, E., Li, B., Rahmati, A., Xiao, C., et al.** 2018. Robust physical-world attacks on deep learning visual classification. In Proceedings of the *IEEE Conference on Computer Vision and Pattern Recognition*, pp.1625-1634.

[4]    **Ferdowsi, A., and Saad, W.** 2019. Deep Learning for Signal Authentication and Security in Massive Internet-of-Things Systems. *IEEE Transactions on Communications*, 67(2), pp.1371-1387.

[5]    **Foot, P.** 1967. The problem of abortion and the doctrine of double effect, *Oxford Review*, 2(2), pp.152-161.

[6]    **Hevelke, A., and Nida-Rümelin, J.** 2015. Responsibility for Crashes of Autonomous Vehicles: An Ethical Analysis." *Science and Engineering Ethics*, 21(3), pp.619-630.

[7]    **Shariff, A., Bonnefon, J. F., and Rahwan, I.** 2017, Psychological roadblocks to the adoption of self-driving vehicles, *Nature Human Behavior*, 1.10, p.694.

[8]    **Thomson, J. J.** 1976. Killing, letting die, and the trolley problem, *The Monist*, 59(2), pp.204-217.

[9]    **Jiang Su** 2018, The Challenges of Self-driving Cars to the Law, *China Law Review*, (2), pp.180-189.

[10]   **Si Xiao, Cao Jianfeng** 2017, On the Civil Liability of Artificial Intelligence, *Science of Law (Journal of Northwest University of Political Science and Law)*, 35(5), pp.166-173.

[11]   **Long Min** 2018. Assignment of the Criminal Liability in Automatic Driving's Accidents, *ECUPL Journal*, 21(06), pp.78-83.

[12]   **Cheng Long** 2018, The Criminal Regulation of Traffic Accidents Caused by Autonomous Vehicles, *Academic Exchange*, No.289(04), pp.83-89

[13]  **Chen Xiaolin** 2016, Challenge and response of self-driving cars to existing laws, Journal of Socialist Theory Guide, 1.

[14]  **Chen Hui, Xu Jianbo** 2014, "The future development trend of smart cars", China Integrated Circuit,11.

[15]  **Gao Zhaoming, Gao Hao** 2017, Information security risk prevention and the value principle of algorithm rule. Philosophical Trends. 09.

[16]  **Li Yiran, Zhou Jie** 2012, Study on active disturbance rejection control for the steering system of the autonomous vehicle, *Journal of Shanghai Normal University(Natural Sciences)*, 2, pp.156-160.

[17]  **Xie, C., Wang, J., Zhang, Z., Ren, Z., and Yuille, A.** 2017, Mitigating adversarial effects through randomization. *arXiv preprint*, arXiv:1711.01991.

[18]  **Xu, Weilin, David Evans, and Yanjun Qi** 2017, Feature squeezing: Detecting adversarial examples in deep neural networks. *arXiv preprint*, arXiv:1704.01155 (2017).